\documentclass[cameraready]{Interspeech}

\title{SDTalk: Structured Facial Priors and Dual-Branch Motion Fields for Generalizable Gaussian Talking Head Synthesis}

\author[affiliation={1}, equalcontribution]{Peng}{Jia}
\author[affiliation={1}, equalcontribution]{Zhen}{Xiao}
\author[affiliation={1}, correspondingauthor]{Jia}{Li}
\author[affiliation={1}]{Xueliang}{Liu}
\author[affiliation={1}]{Zhenzhen}{Hu}
\author[affiliation={2}]{Lingyun}{Yu}

\address{
    $^1$ Hefei University of Technology, Hefei, China \\
    $^2$ University of Science and Technology of China, Hefei, China
}

\email{jiali@hfut.edu.cn}


\usepackage{comment}

\begin{document}

\maketitle

\begin{abstract}
    High-quality, real-time talking head synthesis remains a fundamental challenge in computer vision. Existing reconstruction- and rendering-based methods typically rely on identity-specific models, limiting cross-identity generalization. To address this issue, we propose SDTalk, a one-shot 3D Gaussian Splatting (3DGS)-based framework that generalizes to unseen identities without personalized training or fine-tuning. Our framework comprises two modules with a two-stage training strategy. In the first stage, we incorporate structured facial priors into the reconstruction module and separately predict 3DGS parameters for visible and occluded regions, enabling complete head reconstruction from a single image. In the second stage, we introduce a dual-branch motion field to model coarse and fine facial dynamics, improving detail fidelity and lip synchronization. Experiments demonstrate that SDTalk surpasses existing methods in both visual quality and inference efficiency. The code is available at \href{https://github.com/MSA-LMC/SDGTalk}{https://github.com/MSA-LMC/SDGTalk}.
\end{abstract}

\section{Introduction}
\label{sec:intro}

Talking head synthesis has broad application prospects in film production, human-computer interaction, and virtual reality. This task is to synthesize realistic, temporally coherent facial videos conditioned on driving audio or other control signals.

Existing methods can be categorized into identity-agnostic and identity-specific paradigms according to whether they rely on target identity information. The identity-agnostic paradigm focuses on learning the mapping from audio to facial motion, commonly adopting generative architectures such as GANs~\cite{10.1145/3394171.3413532} and Diffusion Models~\cite{shen2023difftalk}, and training on large-scale, multi-identity video datasets to improve generalization to new identities. These methods~\cite{10.1145/3394171.3413532,shen2023difftalk} emphasize lip synchronization and cross-identity inference, but generally suffer from limited output image quality, difficulty preserving identity traits, and low generation efficiency.

\begin{figure*}[!t]
  \includegraphics[width=1 \linewidth]{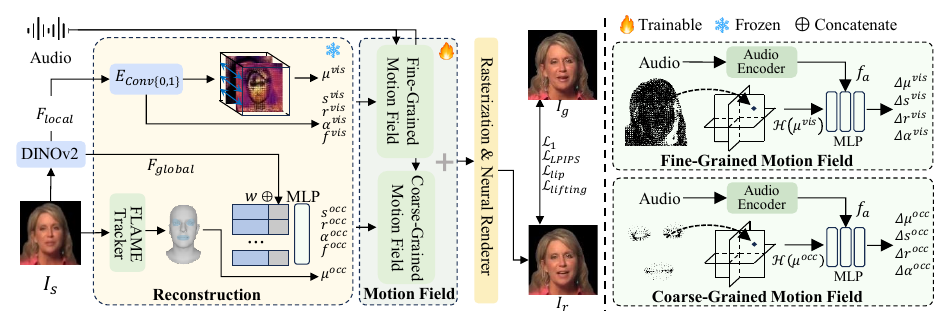}
\caption{\textbf{Overview of SDTalk.} Our approach consists of two main modules: a reconstruction module and a motion field module. Given a source image $I_{s}$ and its corresponding audio, the reconstruction module leverages a learned structured facial prior to obtain the Gaussian head. The motion field then predicts point-wise deformations conditioned on the audio features $f_{a}$. Finally, a rasterizer and a neural renderer synthesizes the final talking head $I_{r}$. During inference, the reconstruction module is executed only once, after which audio-driven animation is achieved solely through the motion field module.}
\label{fig:model}
\end{figure*}

By contrast, identity-specific approaches are based on 3D reconstruction techniques represented by Neural Radiance Fields (NeRF)~\cite{mildenhall2021nerf} and 3D Gaussian Splatting (3DGS)~\cite{kerbl20233d}. NeRF-based methods~\cite{guo2021ad,liu2025nerf} generally require long optimization due to the expensive cost of vanilla NeRF. More recent research~\cite{li2024talkinggaussian,deng2025degstalk} leverages 3DGS to replace implicit volumetric representations with explicit Gaussian models, substantially improving training speed and rendering efficiency; such approaches~\cite{li2024talkinggaussian,cho2024gaussiantalker} can reduce training time to the hour level while maintaining high-quality synthesis and achieving real-time inference performance.

Although reconstruction and rendering (RAR)-based methods better preserve target identity characteristics and thereby mitigate some limitations of generative model-based approaches, they still face significant challenges. On the one hand, acquiring high-quality training video data for specific identities is difficult; on the other hand, training identity-specific models from scratch for new identities often requires substantial time. To overcome these challenges, some works~\cite{ye2024mimictalk,li2025instag} adopt a pretraining and adaptation (PAA) paradigm: the pretraining stage learns universal motion priors from multi-identity data, and the adaptation stage fine-tunes the model to new identities. This strategy partially reduces adaptation time and decreases the required video samples to only a few seconds, but still requires a dedicated fine-tuning step for each new identity.

In this paper, we propose SDTalk, a one-shot talking head generation method based on 3DGS that generalizes to new identities without personalized fine-tuning. As shown in Fig.~\ref{fig:model}, the framework comprises two main components: a reconstruction module and a motion field module. First, due to insufficient geometric information from a single image, direct reconstruction often results in structural deficiencies, making it difficult to recover internal details such as teeth. To address this, we introduce structured facial priors in the reconstruction module, leveraging the geometric priors of the FLAME~\cite{li2017learning} mesh to predict Gaussian parameters for occluded facial regions, while employing a dual-lifting method~\cite{chu2024generalizable} to precisely fit geometry and texture for visible regions, thereby reconstructing a complete Gaussian head. Second, we design dual-branch motion fields: a coarse-grained branch captures global facial motion trends, while a fine-grained branch models local high-frequency details. This design improves lip synchronization and visual fidelity.

\section{Proposed Method}

\subsection{Preliminaries}




3D Gaussian Splatting (3DGS)~\cite{kerbl20233d} represents a 3D scene as an explicit collection of anisotropic Gaussian primitives. Each Gaussian primitive is parameterized by a set of attributes, including a 3D mean $\mu \in \mathbb{R}^3$, a scale vector $s \in \mathbb{R}^3$, a rotation quaternion $r \in \mathbb{R}^4$, an opacity scalar $\alpha \in \mathbb{R}$, and a $Z$-dimensional color feature vector $f \in \mathbb{R}^Z$. Accordingly, a Gaussian primitive $\mathcal{G}$ is defined as:
\begin{equation}
\mathcal{G}=\{\mu, s, r, \alpha, f\},
\end{equation}

Each Gaussian primitive defines a spatial basis function of the form:
\begin{equation}
G(\mathbf{x}) = e^{-\frac{1}{2} (x - \mu)^T \Sigma^{-1} (x - \mu)},
\end{equation}
where $x\in\mathbb{R}^3$ denotes a 3D point in space. The covariance matrix $\Sigma$ is constructed from the scale vector $s$ and the rotation quaternion $r$, encoding the anisotropic shape and orientation of the Gaussian.


\subsection{Structured Facial Priors}

Since a static facial image contains limited information, directly reconstructing a Gaussian head from it often results in structural deficiencies. For example, when the mouth is closed in the input image, the reconstruction typically lacks Gaussian primitives corresponding to internal details such as teeth and the oral cavity. To address this issue, we introduce structured facial priors into the reconstruction module and employ two complementary branches to reconstruct an animatable Gaussian head for visible and occluded regions: i) a visible branch that precisely fits observable surface geometry and texture, and ii) a completion branch that leverages the priors to predict internal structures. We employ DINOv2~\cite{oquab2023dinov2} to extract local features $F_{local}$ and global features $F_{global}$ from the source image $I_{s}$, which serve as conditional information for the two branches, respectively.

\textbf{Visible branch.} For the visible facial regions, we follow prior work~\cite{chu2024generalizable} and adopt a dual-lifting method for reconstruction. Specifically, we initialize a feature plane in 3D space and determine the coordinates $p_{s}$ and normal vectors $n_{s}$ for each point on the plane. Convolutional networks $E_{Conv\{0,1\}}$ are then used to predict the pixel-wise offsets relative to the feature plane as well as the Gaussian parameters excluding position, denoted as $G_{s, r, \alpha, f}^{vis}$. Based on the predicted offsets, positions, and normal vector, we compute the positions of 3D Gaussians $\mathcal{G}_{\mu}^{vis}$. This process can be formulated as:
\begin{equation}
\begin{aligned}
\mathcal{G}_{\mu}^{vis}
&= [p_s \pm E_{Conv\{0,1\}}(F_{local})\cdot n_s] \\
&= \{ \mu^{vis} \},
\end{aligned}
\end{equation}
\begin{equation}
\begin{aligned}
\mathcal{G}_{s, r, \alpha, f}^{vis}
&= [E_{Conv0}(F_{local}),E_{Conv1}(F_{local})] \\
&= \{ s^{vis}, r^{vis}, \alpha^{vis}, f^{vis} \}.
\end{aligned}
\end{equation}

\textbf{Completion branch.} For occluded facial regions, directly regressing the Gaussian parameters is challenging; therefore, we leverage the geometric priors of the FLAME~\cite{li2017learning} mesh. We utilize the FLAME mesh of the source image and select the vertices in the mouth and eye regions as the candidate set for completion. Specifically, each vertex $v_i$ is assigned a learnable weight $w_i$ and concatenated with the global feature $F_{global}$. An MLP then predicts Gaussian parameters excluding position, and use the position of the FLAME mesh vertices $\mathcal{G}_{\mu}^{occ} = \{ \mu^{occ} \}$. This process can be expressed as:
\begin{equation}
\begin{aligned}
\mathcal{G}_{s, r, \alpha, f}^{occ}
&= \mathrm{MLP}(w \oplus F_{global}) \\
&= \{ s^{occ}, r^{occ}, \alpha^{occ}, f^{occ} \}.
\end{aligned}
\end{equation}
where $\oplus$ denotes the concatenation operation.


Finally, the Gaussian primitives generated from the visible and occluded pathways are merged to reconstruct the complete Gaussian head $\mathcal{G} = \mathcal{G}^{vis} \cup \mathcal{G}^{occ}$. Following~\cite{chu2024generalizable}, we predict 32-dimensional color embeddings and synthesize the final image using a neural renderer.

\subsection{Dual-Branch Motion Fields}


In audio-driven facial animation, the core challenge is learning a robust mapping from speech to lip motion. Existing 3DGS-based methods~\cite{li2025instag} are trained on limited identities, restricting their ability to capture diverse speech–lip correspondences and generalize effectively. We instead leverage large-scale multi-identity data to learn cross-identity mappings, improving robustness and lip synchronization.

Unlike prior work~\cite{cho2024gaussiantalker} that represents the entire head with a single Gaussian field, our method adopts a decoupled design. The visible branch produces a dense head representation $\mathcal{G}^{vis}$ (175,232 primitives), while the completion branch $\mathcal{G}^{occ}$ contains only 1,920 primitives and focuses on eye and mouth regions. To address this structural asymmetry, we introduce a dual-branch motion field, consisting of a coarse field for global dynamics and a fine field for local high-frequency details.

Both branches share the same network architecture. In the following, we use the coarse-grained motion field as an illustrative example. Our framework is similar to prior 3DGS-based methods~\cite{li2024talkinggaussian}. Specifically, the positions $\mu$ are first encoded using a tri-plane hash encoder $\mathcal{H}$ and then concatenated with the audio feature $f_a$. The representation is decoded by an MLP to predict point-wise deformation $\delta_D$:
\begin{equation}
\delta_D = \mathrm{MLP}(\mathcal{H}(\mu) \oplus f_a) = \{ \Delta \mu, \Delta s, \Delta r , \Delta \alpha \},
\end{equation}
where $\Delta \mu$, $\Delta s$, $\Delta r$, and $\Delta \alpha$ represent the deformation offsets for position, scale, rotation, and opacity, respectively. Finally, the Gaussian primitive parameters are updated as:
\begin{equation}
\mathcal{G} = \{\mu + \Delta \mu, s + \Delta s, r + \Delta r, \alpha + \Delta \alpha, f \}.
\end{equation}

By leveraging hierarchical motion decomposition, our method ensures stable global dynamics while enhancing the fidelity of local details, thereby producing more natural, fine-grained, and robust audio-driven facial animation.

\subsection{Training Details}

We adopt a two-stage training strategy to first establish stable facial representations and then learn motion fields for expressive talking head synthesis.

\textbf{Stage 1. }
In this stage, the framework does not include the motion field module. All modules are trained except the frozen DINOv2~\cite{oquab2023dinov2}. The completion branch uses the FLAME mesh fitted to the ground-truth image $I_g$. We employ an L1 loss $\mathcal{L}_{1}$ and an LPIPS loss $\mathcal{L}_{LPIPS}$ between the rendered image $I_r$ and the ground-truth image $I_g$ to supervise appearance reconstruction. In addition, following prior work~\cite{chu2024generalizable}, we incorporate a lifting distance loss:
\begin{equation}
\begin{aligned}
\mathcal{L}_{lifting} 
&= \Vert P_{FLAME} - P_{vis} \Vert, \\
P_{vis} 
&= \left\{ 
\arg\min_{\mu^{vis} \in \mathcal{G}^{vis}}
\Vert p - \mu^{vis} \Vert
\,\middle|\, p \in P_{FLAME}
\right\},
\end{aligned}
\end{equation}
where $P_{FLAME}$ denotes the FLAME mesh vertices fitted from the ground-truth image, $P_{vis}$ represents the nearest Gaussian point $\mu^{vis} \in \mathcal{G}^{vis}$ for each vertex $p \in P_{FLAME}$, and $\mathcal{G}^{vis}$ is the set of Gaussian points reconstructed from visible regions of $I_s$.
The overall training objective in Stage 1 is given by:
\begin{equation}
\mathcal{L}_{S1} = \mathcal{L}_{1} + \lambda_{l} \mathcal{L}_{LPIPS} + \lambda_{f} \mathcal{L}_{lifting}.
\end{equation}

\textbf{Stage 2. }
In this stage, parameters from the previous stage are frozen, and only the motion field module is optimized. The completion branch uses the FLAME mesh fitted to the source image. To improve lip synchronization, we apply a lip loss $\mathcal{L}_{lip}$, where the lip region is localized by facial landmarks~\cite{bulat2017far} and optimized using an L1 loss. The training objective in Stage 2 is defined as:
\begin{equation}
\mathcal{L}_{S2} = \mathcal{L}_{1} + \lambda_{l} \mathcal{L}_{LPIPS} + \lambda_{f} \mathcal{L}_{lifting} + \lambda_{p} \mathcal{L}_{lip},
\end{equation}
where $\lambda_{l}$, $\lambda_{f}$, and $\lambda_{p}$ are the weights to balance the losses.

\begin{figure}[!t]
  \includegraphics[width=1 \linewidth]{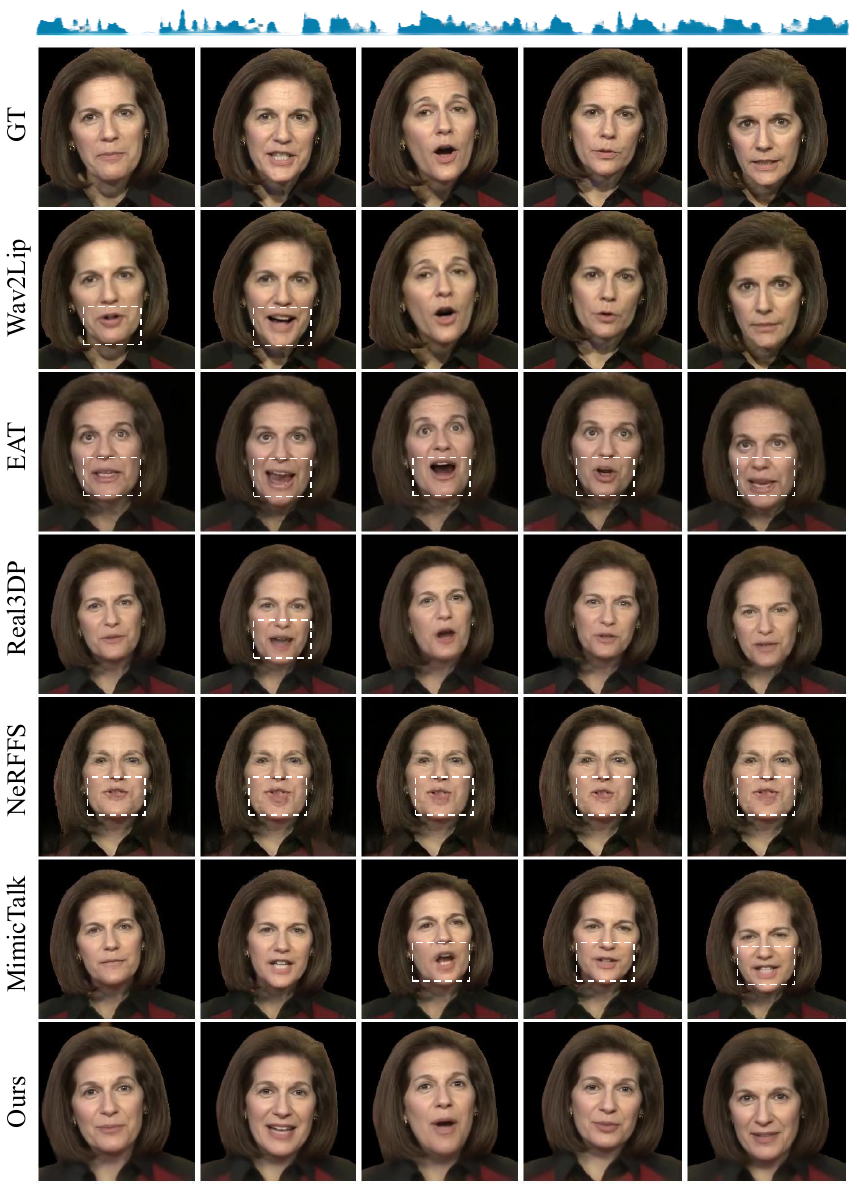}
\caption{Qualitative comparison across different methods. ``Real3DP" and ``NeRFFS" refer to Real3DPortrait \cite{ye2024real3d} and NeRFFaceSpeech \cite{kim2024nerffacespeech}, respectively. Key regions are highlighted by boxes.}
\label{fig:compare}
\end{figure}

\section{Experiments and Results Analysis}

\subsection{Experimental Settings}


\textbf{Dataset. }We train our model on the HDTF~\cite{zhang2021flow} dataset, all videos are cropped and resized to $512\times512$, with camera poses and FLAME parameters estimated using a facial-tracking pipeline and background regions removed. For evaluation, we follow a cross-identity protocol in which test identities are strictly excluded from the training set. In addition, we evaluate our method on the MEAD~\cite{wang2020mead} dataset to further validate its generalization performance across datasets.

\textbf{Implementation Details. }Our method is implemented in PyTorch. Both stages were optimized using the Adam optimizer with a learning rate of $1.0\times10^{-4}$, and each stage was trained for 200,000 iterations. Experiments were run on a single NVIDIA RTX 4090 GPU, and the entire training procedure took approximately 32 GPU hours. Audio features are extracted using HuBERT model~\cite{hsu2021hubert}.

\textbf{Baselines. }
We compare our approach to methods from several paradigms. For zero-shot methods, we include Wav2Lip~\cite{10.1145/3394171.3413532}. For one-shot methods, we compare to EAT~\cite{Gan_2023_ICCV}, Real3DPortrait~\cite{ye2024real3d}, and NeRFFaceSpeech~\cite{kim2024nerffacespeech}. We also include MimicTalk~\cite{ye2024mimictalk}, which follows PAA paradigm.

\textbf{Evaluation Metrics. }
We assess image quality using Peak Signal-to-Noise Ratio (PSNR), Learned Perceptual Image Patch Similarity (LPIPS)~\cite{zhang2018unreasonable}, and Structural Similarity Index (SSIM). Lip synchronization is evaluated using LMD~\cite{chen2018lip} and the SyncNet confidence score (Sync-C)~\cite{chung2016lip}.

\subsection{Quantitative Results}


\begin{figure}[!t]
  \includegraphics[width=1 \linewidth]{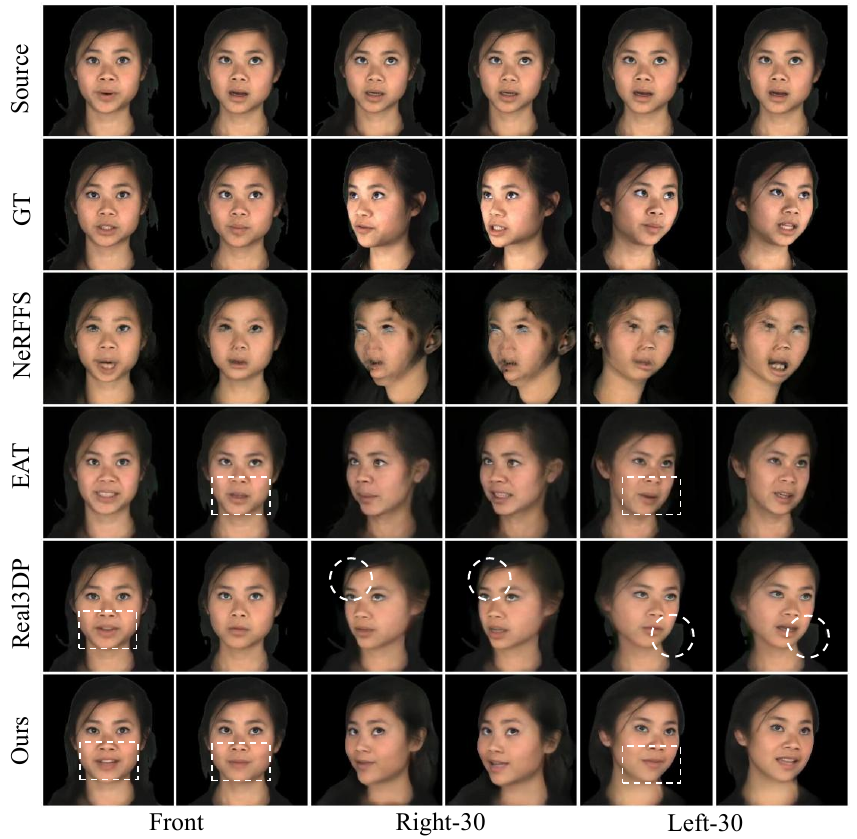}
\caption{Qualitative comparison of talking head synthesis across different methods on the MEAD~\cite{wang2020mead} dataset.}
\label{fig:compare_mead}
\end{figure}

\begin{table}[!t]
\centering
\small
\setlength{\tabcolsep}{1pt}
\caption{Quantitative comparison on the HDTF~\cite{zhang2021flow} dataset. ``ZS'', ``OS'', and ``PAA'' denote \textit{zero-shot}, \textit{one-shot}, and \textit{pretraining-and-adaptation}, respectively. The best results are highlighted in \textbf{bold}, and the second best are \underline{underlined}.}
\resizebox{1\linewidth}{!}{
\begin{tabular}{ccccccccc}
\hline
Methods & Config & PSNR$\uparrow$ & LPIPS$\downarrow$ & SSIM$\uparrow$ & LMD$\downarrow$ & Sync-C$\uparrow$ & FPS$\uparrow$\\ \hline
Wav2Lip~\cite{10.1145/3394171.3413532} & ZS & - & - & - & \textbf{2.40} & \textbf{8.58} & \underline{23} \\ 
EAT~\cite{Gan_2023_ICCV} & OS & 19.22 & 0.31 & 0.55 & 10.83 & \underline{7.91} & 8  \\ 
Real3DP~\cite{ye2024real3d} & OS & 21.51 & 0.21 & 0.72 & 4.61 & 6.70 & 7  \\
NeRFFS~\cite{kim2024nerffacespeech} & OS & 19.15 & 0.26 & 0.61 & 18.68 & 3.46 & 1 \\ 
MimicTalk~\cite{ye2024mimictalk} & PAA & \underline{22.43} & \underline{0.14} & \underline{0.75} & 4.38 & 5.45 & 19 \\ 
Ours & OS & \textbf{24.03} & \textbf{0.13} & \textbf{0.80} & \underline{3.47} & 6.85 & \textbf{45} \\ 
\hline
\end{tabular}
}
\label{tab:quantitative}
\end{table}


\begin{table}[!t]
\centering
\setlength{\tabcolsep}{3pt}
\caption{Quantitative results under fixed head pose.}
\begin{tabular}{cccccc}
\hline
Methods & {PSNR$\uparrow$} & {LPIPS$\downarrow$} & {SSIM$\uparrow$} & {LMD$\downarrow$} & {Sync-C$\uparrow$} \\
\hline
NeRFFS~\cite{kim2024nerffacespeech} & 26.26 & \underline{0.13} & 0.80 & 5.01 & 4.83 \\
EAT~\cite{Gan_2023_ICCV} & 25.59 & 0.19 & 0.77  & 4.33 & \textbf{7.72} \\
Real3DP~\cite{ye2024real3d} & \underline{26.43} & \underline{0.13} & \underline{0.85} & \textbf{2.68} & \underline{7.62} \\
Ours & \textbf{27.78} & \textbf{0.08} & \textbf{0.90} & \underline{3.34} & 6.81 \\
\hline
\end{tabular}
\label{tab:compare_mead_res}
\end{table}


\begin{table}[!t]
\centering
\setlength{\tabcolsep}{3pt}
\caption{Quantitative results under diverse head poses.}
\begin{tabular}{cccccc}
\hline
Methods & {PSNR$\uparrow$} & {LPIPS$\downarrow$} & {SSIM$\uparrow$} & {LMD$\downarrow$} & {Sync-C$\uparrow$} \\
\hline
NeRFFS~\cite{kim2024nerffacespeech} & 14.65 & 0.37 & 0.56 & 26.61 & 5.27 \\
EAT~\cite{Gan_2023_ICCV} & 19.85 & \underline{0.28} & 0.63  & 9.61 & \textbf{7.19} \\
Real3DP~\cite{ye2024real3d} & \underline{19.91} & 0.29 & \underline{0.68} & \underline{3.25} & \underline{6.78} \\
Ours & \textbf{21.17} & \textbf{0.20} & \textbf{0.75} & \textbf{2.83} & 5.41 \\
\hline
\end{tabular}
\label{tab:compare_mead_pose}
\end{table}

Table~\ref{tab:quantitative} reports the quantitative evaluation results on the HDTF~\cite{zhang2021flow} dataset. In general, our method outperforms other approaches in all image quality metrics. Specifically, although EAT~\cite{Gan_2023_ICCV} and Real3DPortrait~\cite{ye2024real3d} achieve relatively high scores on lip synchronization metrics, they are constrained by limited visual fidelity and lower inference speed, making it difficult to simultaneously satisfy both high-quality and real-time requirements. MimicTalk \cite{ye2024mimictalk} relies on identity-specific fine-tuning, which limits its generalization and ease of deployment. By contrast, our method enables real-time inference and achieves the most balanced performance across all evaluation metrics. Tables~\ref{tab:compare_mead_res} and~\ref{tab:compare_mead_pose} present quantitative comparisons on the MEAD~\cite{wang2020mead} dataset compared with other one-shot methods. Under different head poses, our approach consistently outperforms competing models in terms of image quality, demonstrating its strong generalization ability and robustness across various scenarios.



\subsection{Qualitative Evaluation}

We qualitatively compare synthesized sequences across methods in terms of visual fidelity and lip synchronization (Fig.~\ref{fig:compare}, Fig.~\ref{fig:compare_mead}). Wav2Lip~\cite{10.1145/3394171.3413532}, which performs localized mouth-region synthesis, often introduces boundary discontinuities and seam artifacts, compromising global facial consistency. NeRFFaceSpeech~\cite{kim2024nerffacespeech} suffers from accumulated inversion errors, leading to unstable reconstructions. EAT~\cite{Gan_2023_ICCV} and Real3DPortrait~\cite{ye2024real3d} fail to preserve high-frequency facial details, resulting in over-smoothed appearances. In contrast, our method generates more coherent results with better structural and textural preservation. Under large pose variations, although all methods degrade, ours maintains clearer contours and more consistent lip motion, demonstrating superior robustness.

\subsection{Ablation Study}


\begin{table}[!t]
\centering
\setlength{\tabcolsep}{3pt}
\caption{Ablation study results. ``CB'' and ``FM'' denote the completion branch and the fine-grained motion field.}
\begin{tabular}{cccccc}
\hline
Methods & PSNR$\uparrow$ & LPIPS$\downarrow$ & SSIM$\uparrow$ & LMD$\downarrow$ & Sync-C$\uparrow$ \\
\hline
w/o CB & 22.99 & \underline{0.15} & \underline{0.78} & \textbf{3.31} & 5.79 \\
w/o FM & \underline{23.81} & \textbf{0.13} & \textbf{0.80} & 3.52 & \underline{6.48} \\
Ours & \textbf{24.03} & \textbf{0.13} & \textbf{0.80} & \underline{3.47} & \textbf{6.85} \\
\hline
\end{tabular}
\label{tab:ablation}
\end{table}

\begin{figure}[!t]
  \includegraphics[width=1 \linewidth]{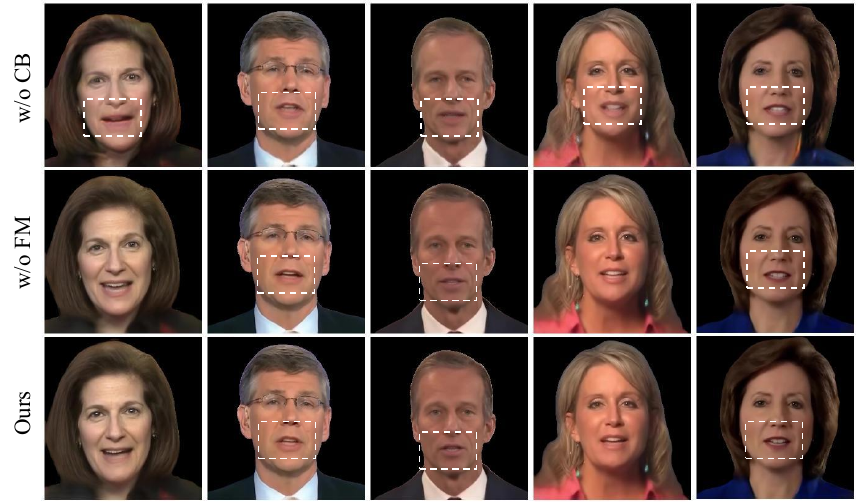}
\caption{Visualization results of the ablation study. Key regions are highlighted by dashed bounding boxes.}
\label{fig:ablation}
\end{figure}

We perform ablation studies to evaluate each component, with quantitative results in Table~\ref{tab:ablation} and qualitative comparisons in Fig.~\ref{fig:ablation}. Removing the completion branch degrades mouth-region quality due to the limited information from a single reference image, hindering the recovery of fine details such as teeth and the oral cavity. Using only the coarse motion field further reduces visual quality and lip synchronization accuracy, highlighting the necessity of the dual-branch motion field design.


\section{Conclusions}

In this paper, we propose a novel one-shot 3DGS-based talking head synthesis method, SDTalk, which significantly improves both visual quality and inference speed. The method addresses the limited information in a single reference image by reconstructing the visible and occluded facial regions separately, and enhances motion modeling by employing dual-branch motion fields that hierarchically decompose facial dynamics. The effectiveness of the proposed method is validated through both quantitative and qualitative experimental results.


\section{Generative AI Use Disclosure}

Generative AI tools were used solely for language editing and polishing of this manuscript. 
All research ideas, methodological design, experiments, and analysis were conducted by the authors. 
No generative AI tools were used to produce the scientific content, experimental results, or core technical contributions of this work. 
All (co-)authors take full responsibility for the content of the paper and consent to its submission.

\bibliographystyle{IEEEtran}
\bibliography{mybib}

\end{document}